\documentclass[submission,copyright,creativecommons]{eptcs}
\providecommand{\event}{ICLP 2019} 
\usepackage{breakurl}             
\usepackage{underscore}           

\usepackage{graphicx}
\usepackage{subfigure}
\usepackage{listings}

\usepackage{tikz}
\usetikzlibrary{arrows.meta}

\title{Encoding Selection for Solving Hamiltonian Cycle Problems with ASP}
\author{
Liu Liu \qquad\qquad Miroslaw Truszczynski
\institute{Department of Computer Science\\
University of Kentucky\\
Lexington, KY 40506, USA}
\email{\quad liu.liu@uky.edu \quad\qquad mirek@uky.edu}
}
\def\titlerunning{ASP encoding selection}
\def\authorrunning{L. Liu and M. Truszczynski}
\begin{document}
\maketitle

 \begin{abstract}
It is common for search and optimization problems to have alternative equivalent encodings in ASP.
Typically none of them is uniformly better
than others when evaluated on broad classes of problem instances.
We claim that one can improve the solving ability of ASP by using machine
learning techniques to select encodings likely to perform well on
a given instance. We substantiate this claim by studying the hamiltonian
cycle problem. We propose several equivalent encodings of the problem and
several classes of hard instances. We build models to predict the behavior
of each encoding, and then show that selecting encodings for a given instance
using the learned performance predictors leads to significant performance gains.
\end{abstract}

\section{Introduction}
Answer Set Programming (ASP) \cite{BrewkaET11} has been shown to be especially
effective on
search and optimization problems whose decision versions are in the class NP,
including many problems of practical interest \cite{GebserMR17,ErdemGL16}.
Despite the ease of modeling and the demonstrated potential of ASP, using it poses
challenges. In particular, it is unlikely a single solver will emerge that
would uniformly outperform other solvers. Consequently, selecting a solver
for an instance may mean the difference between solving the problem within
an acceptable time and having the solver run ``forever.'' To address the
problem, solver
selection, portfolio solving, and automated solver parameter configuration
have all been extensively studied \cite{Rice76,GomesS01,KerschkeHNT19,MarateaPR14,HoosLS14}.
The key idea has been to learn
instance-driven performance models and use them, given an instance, to
select a solver (or a parameter configuration) that might perform well
on that instance.

Another challenge is selecting the right encoding. It is well known that
problems have alternative equivalent encodings as answer set programs. The
problem is that these encodings typically perform differently when run on
different instances. Consequently, one can seek program rewriting heuristics to generate better
performing programs, or develop methods for encoding selection and encoding
portfolio solving, similar to those used in portfolio solving. The first
idea has received some attention in recent years
\cite{BuddenhagenL15,BichlerMW16,HippenL19}.
However, the approach to capitalize on the availability of collections of
equivalent encodings has not yet been explored.

We pursue here this latter possibility and offer for it a proof of concept.
To this end, we study a computationally hard \emph{hamiltonian cycle}
(HC) problem. We construct several ASP encodings of
the problem as well as a collection of hard instances. We show that using
standard machine learning approaches one can build a performance model for
each encoding based on its performance data, and that these performance models
are effective in guiding a selection of encodings to be used with a particular
instance. Our experiments show performance improvements and suggest encoding
selection as a technique of improving ASP potential to solve hard problems.

\section{Encoding Candidates for the HC Problem}
The HC problem has directed graphs as input instances. An instance is given
by the lists of nodes and links (edges), represented as ground atoms over
a unary predicate $\tt{node}$ and a binary predicate $\tt{link}$. The code
below represents a directed graph with four nodes and six links.
{\footnotesize \tt{
\begin{lstlisting}
node(1..4).
link(1,2).link(1,3).link(2,1).link(3,4).link(4,2).link(4,3).
\end{lstlisting}}}

The HC problem imposes constraints on a set of edges selected to form a
solution: there has to be exactly one edge leaving each node, exactly one edge
entering each node, and every node must be \emph{reachable} from every other
node by a path of selected edges. These constraints can be modeled by program
rules such as those below.
{\footnotesize \tt{
\begin{lstlisting}
%Select edges
{ hcyc(X,Y) : link(X,Y) }=1 :- node(X).
{ hcyc(X,Y) : link(X,Y) }=1 :- node(Y).
%Define reachability
reach(X,Y) :- hcyc(X,Y).
reach(X,Z) :- reach(X,Y),hcyc(Y,Z).
%Enforce reachability
:- not reach(X,Y),node(X),node(Y).
\end{lstlisting}}}

The first two rules model the constraints on the number of selected edges
(represented by a binary predicate {\tt hcyc}) leaving and entering each node.
The third and the fourth rule together define the concept of reachability by
means of selected edges. Finally, the last rule, a constraint, guarantees that
every node is reachable from every other node by means of selected edges only.

We can rewrite this encoding to generate its variants. For example,
reachability can be modeled by selecting a node, say 1, and requiring
that every node in the graph (including 1) is reachable from 1 by a
non-trivial path (at least one edge) of selected edges. Another possibility
is to change the way we select edges by rewriting the first two rules.

To obtain a collection of several high-performing encodings for the HC problem,
we generated 15 encodings based on different constraint representation and
rewriting ideas, as discussed above. We ran these encodings on hard instances
to the HC problem (cf. Section \ref{performancedata}) and selected six
encodings based on (1) the percentage of solved instances, and (2) the number
of instances for which an encoding yields the fastest solve time. We refer to
these encodings as Encoding $1,\ldots, 6$.\footnote{All six encodings,
the data sets and detailed experimental results can be accessed at \url{http://cs.uky.edu/~lli259/encodingselection}.}

Table~\ref{encperform} summarizes the performance data for the six encodings
on 784 hard instances (we comment later on how the instances were generated).
The results on the number of wins, instances solved fastest by the encoding,
show that our encodings have complementary strengths. We observe that the
oracle always selecting the fastest solver solves about 98.0\%
of all instances, an improvement of about 16\% over the best individual
encoding. This indicates that there is much room for intelligent encoding
selection methods to improve the performance of ASP on the HC problem.

\begin{table}[tbp]
\centering
	\caption{Performance of individual encodings and the oracle.
} \label{encperform}
	\footnotesize{
	\begin{tabular}{llll}
		\hline \hline
		Encoding & Solved Percentage\% & Average Solved Runtime & Number of Wins\\
		Encoding 1     & 82.3             & 84.1              & 102                 \\
		Encoding 2     & 71.8            & 46.6             & 126                 \\
		Encoding 3     & 55.3             & 29.7              & 110                 \\
		Encoding 4     & 76.2             & 42.9              & 155                 \\
		Encoding 5     & 55.4             & 31.9              & 120                 \\
		Encoding 6     & 77.4             & 47.7             & 151                 \\
		Oracle   & 98.0            & 22.8              & \\
		\hline \hline
	\end{tabular}}
\centering
\end{table}

\section{Data Collection}
\textbf{Performance data.}\label{performancedata}
A natural and often used class of graphs for experimental studies of algorithm
performance is the class of graphs generated randomly from some distribution
space. We designed a program to generate graphs with $n$ nodes and $e$ edges
at random and searched for areas of hardness. We have not identified any such
area. Even when we considered graphs with thousands of nodes and the number of
edges selected from the \emph{phase transition} range, they could be solved
within 10 seconds.\footnote{As $e$ grows (given a fixed $n$), the likelihood
of the graph having a hamiltonian cycle switches from 0 to 1. The region where
the switch occurs is called the \emph{phase transition}. For many problems,
such as satisfiability of $k$-CNF formulas, this is where hard problems are
located \cite{SelmanML96}.}

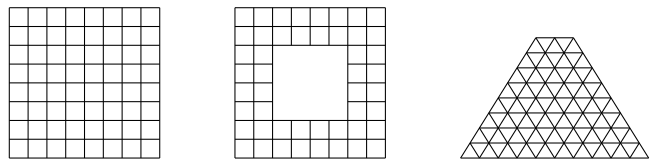
\begin{figure}[!ht]
\centering
\begin{tikzpicture}

\path [-] (0,0) edge (2,0);
\path [-] (0,0.25) edge (2,0.25);
\path [-] (0,0.5) edge (2,0.5);
\path [-] (0,0.75) edge (2,0.75);
\path [-] (0,1) edge (2,1);
\path [-] (0,1.25) edge (2,1.25);
\path [-] (0,1.5) edge (2,1.5);
\path [-] (0,1.75) edge (2,1.75);
\path [-] (0,2) edge (2,2);

\path [-] (0,0) edge (0,2);
\path [-] (0.25,0) edge (0.25,2);
\path [-] (0.5,0) edge (0.5,2);
\path [-] (0.75,0) edge (0.75,2);
\path [-] (1,0) edge (1,2);
\path [-] (1.25,0) edge (1.25,2);
\path [-] (1.5,0) edge (1.5,2);
\path [-] (1.75,0) edge (1.75,2);
\path [-] (2,0) edge (2,2);

\path [-] (3,0) edge (5,0);
\path [-] (3,0.25) edge (5,0.25);
\path [-] (3,0.5) edge (5,0.5);
\path [-] (3,0.75) edge (3.5,0.75);
\path [-] (4.5,0.75) edge (5,0.75);
\path [-] (3,1) edge (3.5,1);
\path [-] (4.5,1) edge (5,1);
\path [-] (3,1.25) edge (3.5,1.25);
\path [-] (4.5,1.25) edge (5,1.25);
\path [-] (3,1.5) edge (5,1.5);
\path [-] (3,1.75) edge (5,1.75);
\path [-] (3,2) edge (5,2);

\path [-] (3,0) edge (3,2);
\path [-] (3.25,0) edge (3.25,2);
\path [-] (3.5,0) edge (3.5,2);
\path [-] (3.75,0) edge (3.75,0.5);
\path [-] (3.75,1.5) edge (3.75,2);
\path [-] (4,0) edge (4,0.5);
\path [-] (4,1.5) edge (4,2);
\path [-] (4.25,0) edge (4.25,0.5);
\path [-] (4.25,1.5) edge (4.25,2);
\path [-] (4.5,0) edge (4.5,2);
\path [-] (4.75,0) edge (4.75,2);
\path [-] (5,0) edge (5,2);

\path [-] (6,0) edge (8.5,0);
\path [-] (6.125,0.2) edge (8.375,0.2);
\path [-] (6.25,0.4) edge (8.25,0.4);
\path [-] (6.375,0.6) edge (8.125,0.6);
\path [-] (6.5,0.8) edge (8,0.8);
\path [-] (6.625,1) edge (7.875,1);
\path [-] (6.75,1.2) edge (7.75,1.2);
\path [-] (6.875,1.4) edge (7.625,1.4);
\path [-] (7,1.6) edge (7.5,1.6);

\path [-] (7,1.6) edge (6,0);
\path [-] (7.25,1.6) edge (6.25,0);
\path [-] (7.5,1.6) edge (6.5,0);
\path [-] (7.625,1.4) edge (6.75,0);
\path [-] (7.75,1.2) edge (7,0);
\path [-] (7.875,1) edge (7.25,0);
\path [-] (8,0.8) edge (7.5,0);
\path [-] (8.125,0.6) edge (7.75,0);
\path [-] (8.25,0.4) edge (8,0);
\path [-] (8.375,0.2) edge (8.25,0);

\path [-] (7.5,1.6) edge (8.5,0);
\path [-] (7.25,1.6) edge (8.25,0);
\path [-] (7,1.6) edge (8,0);
\path [-] (6.875,1.4) edge (7.75,0);
\path [-] (6.75,1.2) edge (7.5,0);
\path [-] (6.625,1) edge (7.25,0);
\path [-] (6.5,0.8) edge (7,0);
\path [-] (6.375,0.6) edge (6.75,0);
\path [-] (6.25,0.4) edge (6.5,0);
\path [-]  (6.125,0.2) edge (6.25,0);
\end{tikzpicture}
\caption{Structured instances: regular grid graph, regular grid graph with
a hole, regular triangular graph with cutting area }\label{fig2}

\end{figure}

To find classes of graphs for which the HC problem is not easily solved in ASP,
we consider structured graphs (structure often is a source of hardness).
Starting with some highly structured graph that has a hamiltonian cycle,
we remove edges at random until the graph is no longer hamiltonian.
In this work, we start the process with grid graphs shown in
Figure~\ref{fig2}. We set their dimensions and ``hole'' locations so that to
guarantee the existence of a hamiltonian cycle. Our experiments demonstrate
that graphs with the number of edges in the phase transition region tend to
yield programs that often require hundreds or thousands of seconds, even
when the graphs have relatively few nodes (of the order of hundreds).

To collect performance data, we generated a large collection of graphs.
Then we combined each graph with each of the six encodings, and ran
\emph{clasp/gringo}\footnote{\url{https://potassco.org}} on the resulting
programs. We set the cutoff time to 200s. When an instance timed out, we
used the penalized runtime as an approximation of its real runtime, computed
based on the number $k$ of encodings for which the instance timed out.
We then selected instances with runtime between 50s and 200s for at least one
encoding (not necessarily the same one). We call these instances
\emph{reasonably hard} (those that cannot be solved with any encoding in under
200s are too hard, and those that can be solved with each encoding in under 50s
are too easy). Finding reasonably hard instances is time
consuming. Our experiments show that only a small fraction falls into this
category (cf. Table \ref{tabledistri}). Thus, typically several graphs need to be
generated before a single reasonably hard instance is found. We used this
method to build a collection of 784 reasonably hard instances.\footnote{The graph instances and the performance data
can be downloaded from \url{http://cs.uky.edu/~lli259/encodingselection}}

%

\begin{table}[tbp]
\caption{Runtime distribution of Encoding 2 on 500 different instances that are gained through eliminating random edges from a regular square grid graph with length of its side 10.}
\label{tabledistri}
\centering
\footnotesize{
\begin{tabular}{llll}
	\hline\hline
runtime & \textless{}50s & 50s $\sim$200s & 200s$\leq$ \\
counts   & 330            & 52             & 118  \\
	\hline\hline
\end{tabular}}
\centering
\end{table}

\smallskip
\noindent
\textbf{Instance features.}
To use machine learning to construct a predictor of performance for a given
instance on a particular encoding, we need informative and easy to compute
instance features. In this work, we considered features of two types: graph
features and encoding-based features. Some graph features capture general
characteristics of graphs such as the numbers of nodes and edges, or the
minimum and the maximum degrees. Other graph features are constructed to
reflect aspects of the problem at hand. In our case, they are designed to
capture properties of depth-first and breadth-first search trees rooted in
nodes of the graph as they inform about reachability from a node.\footnote{All
features and explanations are available at \url{http://cs.uky.edu/~lli259/encodingselection}}

Encoding-based features of an instance are obtained by means of the program
\emph{claspre}\footnote{\url{https://potassco.org/labs/claspre/}}. It extracts
\emph{static} and \emph{dynamic} features of ground ASP programs while solving
them for a short amount of time. In order to obtain \textit{claspre} features
of a graph instance, we combine the instance with our six encodings and then
pass the resulting ground programs to \textit{claspre}.
In total, there are 569 features in 13
groups, one group of graph features and six pairs of groups of \textit{claspre}
encoding-based static and dynamic features. When learning performance
predictors (the details are in the next section), we used a narrowed down set
of features to avoid overfitting and retain features that are informative for
the HC problem.

\section{Encoding Selection with Machine Learning}
The goal of encoding selection is to identify encodings
that promise good performance for a given instance. Our work is based on the performance data and instance features computed and
collected for the data set of 784 reasonably hard instances. We use
this data to build regression models for the six encodings we constructed as
representations of the HC problem. Specifically, we build $k$-nearest neighbor
(KNN), decision tree (DT) and random forest (RF) regressors. All these models
are directly imported from python \emph{scikit-learn} package.

To select informative features 
we perform
in-group individual feature selection followed by the feature group
selection. We start with an empty feature set, randomly add one or more
features, and then test the average performance of selected features to decide whether to
keep them or not.

To evaluate a particular set of selected features, we randomly divide our data
into the training set (80\%) and the test set (20\%), train models using the
training set, and test the performance of encoding selection result on the
test set.
We partition training data into 10 bins and use 10-fold cross-validation
to improve the generalization performance.
The approach we described in this section results in a set of 41 features
(cf. Table~\ref{ecfeatures}), 27 graph features and 14 \emph{claspre} static
features, all obtained with Encoding 1.
Our results show that \emph{clapsre} dynamic features are not as informative as
graph features and \emph{claspre} static features. We note, however, that because we
used the greedy feature selection, our set of features may not be optimal and
better selections may be possible.

\begin{table}[tbp]
\caption{Selected instance features for encoding selection} \label{ecfeatures}
\centering
{\footnotesize

	\begin{tabular}{lll}
		\hline\hline
ratio\_node\_edge                 & avg\_depth\_beam              & Frac\_Binary\_Rules\_hc1          \\
ratio\_bi\_edge                   & dfs\_1st\_back\_depth         & Frac\_Ternary\_Rules\_hc1         \\
avg\_out\_degree                  & sum\_of\_choices\_along\_path & Free\_Problem\_Variables\_hc1     \\
avg\_in\_degree                   & depth\_avg\_dfs\_backjump    & Problem\_Variables\_hc1           \\
ratio\_of\_odd\_out\_degree       & depth\_back\_to\_root         & Assigned\_Problem\_Variables\_hc1 \\
ratio\_of\_even\_out\_degree      & depth\_back\_to\_any          & Constraints\_hc1                  \\
ratio\_of\_odd\_in\_degree        & depth\_one\_path              & Rules\_hc1                        \\
ratio\_of\_even\_in\_degree       & min\_depth\_bfs               & Frac\_Normal\_Rules\_hc1          \\
ratio\_of\_odd\_degree            & max\_depth\_bfs               & Frac\_Cardinality\_Rules\_hc1     \\
ratio\_of\_even\_degree           & avg\_depth\_bfs               & Frac\_Choice\_Rules\_hc1          \\
ratio\_out\_degree\_less\_than\_3 & min\_depth\_beam              &       Frac\_Binary\_Constraints\_hc1     \\
ratio\_in\_degree\_less\_than\_3  & max\_depth\_beam              & Frac\_Ternary\_Constraints\_hc1    \\
ratio\_degree\_less\_than\_3      & avg\_depth\_beam              & Frac\_Other\_Constraints\_hc1   \\
avg\_depth\_bfs                   & Frac\_Unary\_Rules\_hc1      &   \\
\hline\hline
	\end{tabular}
	\centering}
\end{table}

\section{Experimentation}

\textbf{Hardware.}
Our experiments were conducted on a computer with four cores, each with
Intel i7-7700 3.60GHz CPU and 16GB RAM, running under 64-bit 18.04.2 LTS
(Bionic Beaver) Ubuntu system. The solver used is
\emph{clasp}\footnote{\url{https://potassco.org}} version
3.3.2 with default parameter setting. The grounding tool
is \emph{gringo}\footnote{\url{https://potassco.org}} version 4.5.4. We choose
200 CPU seconds as
cutoff time.

\smallskip
\noindent
\textbf{Analysis of encoding selection results.}
We performed encoding selection experiments using different machine learning
methods on the training set and test set described earlier. The models were
trained with the narrowed down set of features and hyper-parameters set to
values obtained via the standard hyper-parameter tuning process.
The results are shown in Table~\ref{ecresult1}.

The test set contains 156 instances randomly selected from the original
data set of 784. The results show that
the encodings we used in the experiment have complementary strengths. The oracle solves 98.7\%, or 154 out of 156 instances.
Compared with Encoding 1, the
always-select-best selection method solves 14.7\% more instances.
We note that 98.7\% is the upper bound for the performance that could
be achieved by encoding selection.

Our best predictor based on the decision tree method, solves 96.2\% of
instances, or 150 out of 156. This is very close to 98.7\% of solved
instances for the always-select-best oracle. Even the worst performing of
three machine learning models, based on the $k$-nearest neighbors algorithm,
solves 92.9\% of instances, much better than any individual encoding. Overall, we
find that each of the three machine learning methods we studied provide
promising results in terms of the percentage of the solved instances.

\begin{table}[tbp]
\caption{Test result of encoding selection experiment } \label{ecresult1}
\centering
\footnotesize{
\begin{tabular}{llll}
\hline\hline
                  & Solved Percentage\% & Average Solved Runtime & Number of Wins \\ \hline%
Single encoding performance & & &                                     \\ %
Encoding 1        & 84.0           & 82.4                      & 25            \\ %
Encoding 2        & 71.2           & 44.0                      & 29            \\ %
Encoding 3        & 56.4           & 30.7                       & 20            \\ %
Encoding 4        & 78.8           & 38.6                      & 28            \\ %
Encoding 5        & 57.1           & 35.4                      & 26            \\ %
Encoding 6        & 79.4           & 48.1                      & 26            \\\hline %
Oracle performance  & & &                                               \\ %
Oracle            & 98.7           & 21.1                      &               \\ \hline
Encoding selection & & &                                                \\ %
Encoding selection (KNN) & 92.9          & 40.2                    &               \\ %
Encoding selection (DT) & 96.2&42.2       &               \\ %
Encoding selection (RF) & 93.6&41.7	      &               \\ %
\hline\hline
\end{tabular}}
\centering
\end{table}

\section{Summary}
We applied machine learning techniques to build performance prediction models
for a collection of six encodings of the HC problem that showed complementary
strengths on our data set. We designed features to characterize problem
instances (some of them problem independent and some reflecting the problem of
interest). Finally, we applied three kinds of regression models to construct
performance predictors. We used these predictors to select an encoding
for an instance to run it on an ASP solver. The results showed performance
gain over individual encodings. Moreover, the encoding selection approach
came very close to the always-select-best oracle in terms of solved instances.
We conclude that the encoding selection method improves the solving
capabilities of ASP solvers.

\section{Future Work}

It is unsatisfactory that the running time of our encoding selection approach
is higher (about two times higher in our experiments) than the optimal time of
the always-select-best oracle. Closing that gap is a challenge that will
require more accurate runtime prediction. One way to attack the problem is
to identify more informative features, especially domain-specific features.
Also, feature selection can remove irrelevant and redundant attributes and has huge impacts on the performance of machine learning models.

Another factor that affects the performance of solving an ASP problem is a
specific solver used to perform the search. The default parameter configuration may be good overall but more
often than not will not be optimal on specific instances. Work on parameter
configuration, such as ParamILS~\cite{HutterHLS09}, has shown that a
well-chosen parameter configuration can help achieve a performance improvement
of over one order of magnitude. Our next step is to combine encoding selection with parameter
configuration.

The encoding candidates in our experiments were created and modified manually.
However, it is desirable to automate the
process, that is, to generate candidates with a tool that is able to analyze
an encoding and rewrite it into several equivalent and efficient forms.

\section*{Acknowledgements}
This work was funded by the NSF under the grant IIS-1707371.

\nocite{*}
\bibliographystyle{eptcs}
\bibliography{refs}
\end{document}